\documentstyle[11pt]{article}

\def\thebibliographya#1{\section*{Other References}\list
 {[\arabic{enumi}]}{\settowidth\labelwidth{[#1]}\leftmargin\labelwidth
 \advance\leftmargin\labelsep
 \usecounter{enumi}
}
 \def\newblock{\hskip .11em plus .33em minus .07em}
 \sloppy\clubpenalty4000\widowpenalty4000
 \sfcode`\.=1000\relax}

\newcounter{max}
\def\thebibliographyb#1{
 \setcounter{max}{\theenumi}
 \list
 {[\arabic{enumi}]}{\settowidth\labelwidth{[#1]}\leftmargin\labelwidth
 \advance\leftmargin\labelsep
 \usecounter{enumi}
}
 \setcounter{enumi}{\themax}
 \def\newblock{\hskip .11em plus .33em minus .07em}
 \sloppy\clubpenalty4000\widowpenalty4000
 \sfcode`\.=1000\relax}

\begin{document}

%% ABSTRACT TO BE SUBMITTED/USED SEPARATELY IN CONFERENCE --
%% DON'T INCLUDE IN MAIN BODY FOR PROCEEDINGS

%% (but INCLUDED in this version!)
%\iffalse

{\Large To appear in {\bf Proceedings of the CBCL Learning Day (1994)}}\\

{\large following Oral Presentation by Gregory Galperin at the CBCL
  Learning Day (Jan. 1994)}\\[8mm]

{\bf Abstract:}
%% KEEP FIRST & LAST SENTENCES IN SYNC WITH "IMPACT" SECTION ABOVE
A novel approach to learning is presented, combining
features of on-line and off-line methods to achieve considerable
performance in the task of learning a backgammon value function in a
process that exploits the processing power of parallel supercomputers.
The off-line methods comprise a set of techniques for parallelizing
neural network training and $TD(\lambda)$ reinforcement learning;
here Monte-Carlo ``Rollouts'' are introduced as a massively parallel
on-line policy improvement technique which applies resources to the
decision points encountered during the search of the game tree
% (rather than spending a given computational budget on achieving
% uniform learning gains across all possible states, most of which
% will never be visited)
%to achieve a more accurate estimate of the value function.
to further augment the learned value function estimate.
A level of play roughly as good as, or possibly better than, the
current champion human and computer backgammon players has been
achieved in a short period of learning.
%% KEEP FIRST & LAST SENTENCES IN SYNC WITH "IMPACT" SECTION ABOVE
%% (should I make thoe into latex macros??  seems like overkill...)

\section*{}\section*{}
\newpage

%\fi

\begin{center}
{\bf LEARNING AND IMPROVING BACKGAMMON STRATEGY}\\[8mm]
GREG GALPERIN\\[8mm]
\end{center}

{\bf The Problem:}
From an applications perspective, the challenge is to design and
implement a system that learns how to play backgammon at a level that
surpasses the best human players in the world.  On the theoretical
side, the goal is to develop fast methods for learning and using
an optimal or near-optimal policy in a high-dimensional,
delayed-reward, stochastic setting.

The task of playing a game well can be reduced to the task of picking
the best move at each decision point, where the best move is the one
that maximizes the probability of winning.  A value function that is
meant to represent the probability of winning can be approximated, and
decisions are still made in the same way using this approximate value
function.  The value function has traditionally been a set of man-made
heuristics that represent human experts' intuitions about the games.
Instead, computers can learn to play board games simply by learning
the evaluation function based on experience in playing or
training from samples of positions taken from games.

% This is a precise concept, since the game tree including all possible
% legal moves (in which nodes are either decision points for you or your
% opponent, or stochastic events) can in theory be completely
% enumerated, and the exact probability of winning at each point
% (assuming two optimal players) computed.  The optimal player would
% then, at each decision point, simply make the move that reached the
% node giving the greatest probability of winning.  Since an exhaustive
% tree search for a game like backgammon is computationally intractable,

{\bf Motivation:}
Perfect information board games are excellent domains for investigating
new learning methodologies; the creation of evaluation functions is
very much a black art, and is a labor intensive process.  Generally,
it is done via a trial and error process through a decade or more of
human effort for each player, and the result is generally a somewhat
cryptic and complicated hand-crafted evaluation function.  Such a
representation of knowledge about the game is certainly not in the
same form as its designers' internal representations, but is an attempt
to recast that knowledge into a form useful to computers; even if that
translation process were perfect with no information lost, that
computer player would still not know any more than its programmers did,
and it is likely that in practice much less information is captured.
Instead, a learning process could learn information relevant to the
chosen representation, would not be constrained by the knowledge of
its ``instructors'' (nor would it benefit from their ability), and
would automate the tedious and mysterious process of developing the
evaluation function.

Second, since the structure of the value function is not known, a
general function approximator must be used to represent the learned
function.  It is almost certain that with a finite time to train the
system (thus requiring a model of finite dimensionality), there will
always be some error in the approximation of the function.  Given a
value function that is known to be imperfect, an appropriate boosting
procedure might improve the performance of the evaluation function,
and would augment the learning process.

% Perfect information board games are excellent arenas for testing new
% learning techniques:
% \begin{itemize}
% \item The structure of the problem is simple, and the rules (state
% transitions and rewards) are well-defined.
% \item Yet, the task is extremely difficult as a whole.
% \item The space of possible states is immense (backgammon has
% about $10^{24}$ possible states), yielding a ``realistically complex''
% problem.
% \item Performance is easy to measure: play games against computers or
% people, and count the results (a tournament!).
% \item Humans intuitively understand these domains, and thus can have
% insight into the computer's progress, oversights, innovations, and 
% representation.
% \end{itemize}

{\bf Previous Work:}
There is a long history of development of computer games, though most
of this work has been focused on hand-crafted evaluation functions
\cite{berliner-bkg} and strategies for tree search and heuristic tree
pruning.  Tesauro initially used human expert rankings of moves to
train an evaluation function \cite{tesauro-neurogammon}, and later
used $TD(\lambda)$ reinforcement learning \cite{sutton-td} to allow a
system to learn the evaluation function through self-play
\cite{tesauro-td-gammon}.  Monte Carlo techniques have been studied
rigorously in many applications outside of learning and games
\cite{monte-carlo}.

{\bf Approach:}
A functional representation (``architecture of the learning box'') is
chosen, and the value function for the game of backgammon is learned
on a parallel computer by examples from self-play.  Very simple
(linear) representations have produced some of the best results.
Then a backgammon game is played against an opponent with the help of
that learned function in the following manner: at each decision point,
all legal actions are considered; for each action, many games are
played to termination in parallel with all decisions being made for
both sides based on the learned value function (in the obvious manner:
the function is evaluated for all possible moves, and the move that
maximizes the value function is taken).
% Even with an imperfect learned
% value function, critically here the {\em same} imperfect player is
% playing both sides; without any bias in the players or in the random
% dice rolls, any difference in outcomes will be due to the inherent
% strength or weakness of the position.
% For efficiency, confidence
% bounds are maintained around each resulting estimate of the value
% of each action, and while an action's upper bound is less than the
% best action's lower bound no games are played for that action.
%% (Note that the bounds are asymmetric; the choice between actions
%% can be modeled as Multi-Armed Bandit.)
The action that led to the
greatest percentage of wins is chosen.

The on-line phase of the algorithm approximates a Monte Carlo
estimation of the probability for an optimal player to win when
playing against another optimal player.  Each game played to the end
(here termed a ``Rollout'' of the position under consideration)
is a random sample from the space of all possible roll sequences; an
exhaustive search over this space is clearly intractable.  This
process is embarrassingly parallel: a massive number of processors can
play games simultaneously starting from the same point to collect
Monte Carlo statistics.  Furthermore, the parallelism is crucial to
making the system run in real time, so that decisions can be made in
tens of seconds rather than in hours.  Applying this Monte Carlo
``Rollout''
technique has achieved improvements to the point that this method
beats the originally learned function 75\% of the time, exhibiting a
greater margin of victory (implying a better job of
learning/representing the optimal value function) than a significantly
more complicated functional representation is able to achieve with
orders of magnitude more initial learning time.
% Given a fixed computational budget such as supercomputer time, our
% approach reallocates off-line (training-time) resources to instead
% be used on-line (performance-time) for a substantial improvement in
% overall task performance.

{\bf Difficulty:} A proof that this Monte Carlo technique achieves any
improvement over the learned function simply used in the obvious
manner is formidable.  For this to be true, the evaluation function
must not be coherently flawed in some manner as to systematically
overlook some important strategy or policy.  Assuming that the error
in the value function is unbiased white noise may allow argument that
the average performance is no worse than the value function, but
stronger results will be difficult.  In practice, this is a naive
assumption, though all results observed have shown strict improvement.
Other issues include the fact that this technique is tailored for
decisionmaking in a stochastic environment.

% There are several assumptions that must be made about the
% structure and performance of the imperfect evaluation function:
% \begin{enumerate}
% \item The evaluation function must not be ``systematically flawed'' in
% some manner.  A mathematical translation of this constraint is that
% the ``error'' in the value function (departure from the ideal/true
% value function) is zero-mean white noise.  Intuitively, this could
% mean that the value function does not overlook a certain class of
% features in determining value.  For instance, an example of a
% ``systematically flawed'' evaluation function might not consider
% making a short-term piece sacrifice for an eventual gain.
% \end{enumerate}

{\bf Impact:}
%% KEEP IN SYNC WITH "ABSTRACT" SECTION ABOVE
This technique proposes a novel approach to learning, combining
features of on-line and off-line methods to achieve considerable
performance in the task of learning a backgammon value function in a
process that exploits the processing power of parallel supercomputers.
A level of play roughly as good as, or possibly better than, the
current champion human and computer backgammon players has been
achieved in a short period of learning.

\vspace{5mm}
REFERENCES

\begin{thebibliographyb}{5}

\bibitem{berliner-bkg} H.J. Berliner, ``BKG---A Program That Plays
Backgammon,'' {\it Computer Games I}, D.N.L. Levy, pp. 3--28 (1988).

\bibitem{tesauro-neurogammon} G. Tesauro, ``Neurogammon: A
Neural-Network Backgammon Program,'' {\it IJCNN Proceedings} {\bf
III}, 33--39 (1990).

\bibitem{sutton-td} R.S. Sutton, ``Learning to Predict by the Methods
of Temporal Differences,'' {\it Machine Learning} {\bf 3}, pp. 9--44
(1988).

\bibitem{tesauro-td-gammon} G. Tesauro, ``Practical Issues in Temporal
Difference Learning,'' IBM Technical Report RC-17223, \#76307 (1991).

\bibitem{monte-carlo} R.M. Neal, ``Probabilistic Inference Using
Markov Chain Monte Carlo Methods,'' University of Toronto Dept. of
Computer Science Technical Report CRG-TR-93-1 (1993).

\end{thebibliographyb}

\end{document}